\PassOptionsToPackage{table}{xcolor}
\documentclass[]{antgroup}
\usepackage{antgroup}

\usepackage{latexsym}

\usepackage[T1]{fontenc}

\usepackage[utf8]{inputenc}

\usepackage{amsmath}
\usepackage{amssymb}
\usepackage{booktabs}
\usepackage{enumitem}
\usepackage{graphicx}
\usepackage{multirow}
\usepackage{xspace}
\usepackage{xcolor}
\usepackage{tikz}
\usepackage{pgfplots}
\usepackage{flafter}
\usepackage{float}
\usepackage{placeins}
\pgfplotsset{compat=1.18}
\usetikzlibrary{arrows.meta,calc,decorations.pathreplacing,positioning,shadows}

\usepackage{algorithm}
\usepackage{algpseudocode}
\usepackage{amsmath,amsfonts,bm}

\def\eqref#1{equation~\ref{#1}}
\def\1{\bm{1}}

\DeclareMathAlphabet{\mathsfit}{\encodingdefault}{\sfdefault}{m}{sl}
\SetMathAlphabet{\mathsfit}{bold}{\encodingdefault}{\sfdefault}{bx}{n}

\definecolor{pcdgreen}{HTML}{2F7D32}
\definecolor{pcdgray}{HTML}{5F6368}
\definecolor{pcdrow}{HTML}{F2F3F5}
\definecolor{figtwoblue}{HTML}{5FA0D6}
\definecolor{figtwoyellow}{HTML}{F3D582}
\definecolor{lladablue}{HTML}{4F7FB8}
\definecolor{lladaorange}{HTML}{F26B3A}

\title{UBTree: Parallel Tree Drafting via Unigram and Bigram Models for Speculative Decoding}

\author{\centering Chumeng Liang$^{1,2,*}$ Linxuan Wang$^{1,3,*}$, Xinyu Peng$^{1,*}$, Huabin Liu$^{1,\dagger}$, Xiaoxu Guo$^{1}$,\\ Yuxin Chen$^{2}$, Ge Liu$^{2}$, Guang Lin$^{3}$, Qifan Song$^{3}$, Jianguo Li$^{1,\dagger}$}
\affiliation{\centering $^1$Inclusion AI \quad\centering $^2$University of Illinois Urbana-Champaign  \quad\centering $^3$Purdue University\\[1.5mm]
\href{https://www.inclusion-ai.org/ubtree/}{\textcolor[cmyk]{0,1,0,0}{\texttt{https://www.inclusion-ai.org/ubtree/}}}}

\begin{document}
\maketitle
\begingroup
\renewcommand{\thefootnote}{}
\footnotetext{\textsuperscript{*}Equal contribution. \textsuperscript{\ensuremath{\dagger}}Corresponding authors.\\
\hspace*{1.8em}This work was done during Chumeng's and Linxuan's internship at Inclusion AI.}
\endgroup

\begin{abstract}
Speculative decoding accelerates language model inference by verifying multiple draft tokens in a single target-model pass. Recent parallel drafters have achieved breakthrough performance in frontier production models, but their effectiveness deteriorates as the entropy of target distributions increases due to insufficient draft diversity. To overcome this bottleneck without sacrificing parallelism, we introduce \textbf{UBTree}, a parallel drafter that couples a \textbf{U}nigram proposer with a \textbf{B}igram selector to construct drafting \textbf{Tree}s. The unigram proposer is trained with the standard cross-entropy objective to generate candidate tokens independently for each position, while a lightweight bigram selector predicts transition scores between adjacent candidate pairs. Unlike the proposer, the selector is trained with a renormalized KL objective on high-temperature data. This \textit{tree-native} training broadens the supervision beyond the greedy path, encouraging plausible alternative branches that improve the chance of accepting additional tokens during tree verification. Across seven standardized benchmarks with Qwen3-4B and Qwen3-8B, UBTree achieves an average speedup of $5.84$--$6.94\times$ over autoregressive decoding and outperforms DARTree in all 28 comparisons. Production-scale evaluation further demonstrates UBTree's advantage over frontier baselines such as DSpark.
\end{abstract}

\begin{figure}[!htbp]
    \centering
    \vspace{-0.2cm}
    \includegraphics[width=\linewidth]{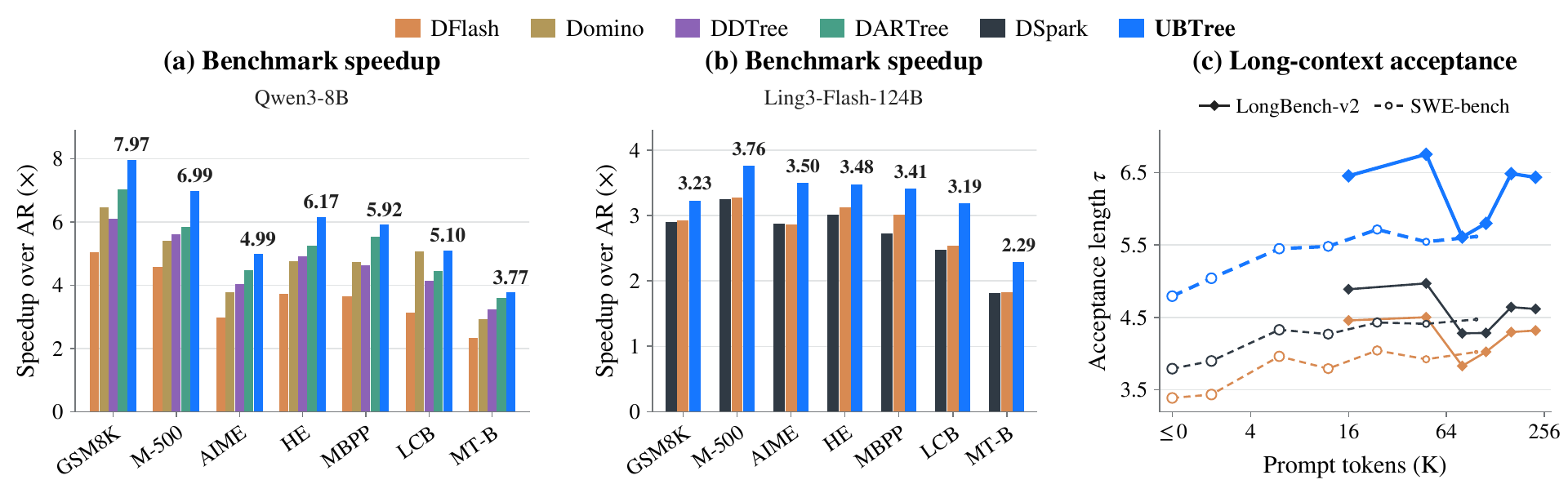}
    \vspace{-0.2cm}
    \caption{\textbf{(a) Speedups on Qwen3-8B:} Speedup over autoregressive decoding across seven benchmarks on Qwen3-8B instruct at $T_{\mathrm{infer}}=1$.
     \textbf{(b) Speedups on Ling3-Flash-124B:} Serving speedup in SGLang on Ling3-Flash-124B at $T_{\mathrm{infer}}=1$.
    \textbf{(c) Long-context acceptance:} Acceptance length $\tau$ on Ling3-Flash-124B for LongBench-v2 and SWE-bench over different context lengths. UBTree outperforms competing methods across most benchmarks and model scales.}
    \label{fig:ubtree-teaser}
\end{figure}

\section{Introduction}
Speculative decoding~\citep{leviathan2023fast,chen2023accelerating} offers a lossless approach to accelerating large language models (LLMs): a lightweight drafter proposes multiple future tokens, and the target model verifies them together while preserving its output distribution. The benefit of speculative decoding depends not only on the acceptance length of the draft by the target model but also on the computational cost. Longer accepted drafts do not necessarily yield greater speedup if producing them incurs substantial latency.

Recent advances in speculative decoding have been driven by breakthroughs in parallel drafting~\citep{liu2026dart,chen2026dflash}. While their autoregressive counterparts~\citep{li2024eagle,li2026eagle} condition each draft token sequentially on preceding choices, parallel drafters predict all future positions in a drafting block with a single proposer-backbone forward pass. Parallelism reduces the sequential overhead of using larger proposer backbones which increases the acceptance length. However, parallel drafters lack draft diversity. Consequently, their performance degrades when high-entropy target distributions inherently produce different paths. Moreover, the missing token-wise dependency may worsen the degradation by mixing these different paths. Existing approaches try to complement this drawback by restoring token-wise dependencies to improve precisions of the single-path draft through lightweight autoregressive correction heads~\citep{huang2026domino,cheng2026dspark} or intra-backbone designs~\citep{inco2026dflash2}. However, these strategies do not fundamentally solve the diversity bottleneck because there is still only one single draft path covered. This particularly limits the further improvement of parallel drafters and their applications under scenarios with high entropy target distributions~\citep{devic2025trace,yang2026longspec}. \textbf{Draft diversity thus becomes a central bottleneck for current parallel speculative decoding.}

A natural solution to this bottleneck is tree drafting~\citep{miao2024specinfer,cai2024medusa,li2024eagle2}, which retains multiple draft paths as tree branches, increasing the chance that target verification matches one of them. By providing diverse yet plausible candidate drafts, tree drafting is more robust in high-entropy scenarios. Combining tree drafting and parallel drafters by constructing budgeted trees from pretrained block-parallel drafters achieves considerable speedups~\citep{ringel2026accelerating,li2026dartree}. Nevertheless, these methods exhibit two limitations. First, the tension between token-wise dependency and full parallelism constrains the tradeoff between draft quality and drafting latency. Existing tree-based methods use autoregressive heads to capture token-wise dependencies, which precludes full parallelization. Conversely, omitting token-wise dependency limits the acceptance length and, consequently, the end-to-end speedup. Second, tree-based parallel methods suffer from a \textit{train--inference mismatch}: they directly apply parallel drafters trained for single-path drafting to multi-path tree verification. Such drafters are optimized to produce the top draft for single-path verification, but are suboptimal for proposing multiple plausible paths for tree verification. Consequently, although lower-ranked alternatives in the draft tree can populate additional branches, they are not sufficiently optimized to align with the target model's alternatives or worth allocating verification budget to.

We address these limitations with \textbf{UBTree}, a tree-based drafter with a parallel bigram selector to model token-wise dependencies and tree-native training to close the train--inference gap. We incorporate two designs in UBTree:
\begin{itemize}
    \item \textbf{Architecture.} UBTree separates drafting into three complementary stages and keeps all neural computation parallel. First, we use DFlash~\citep{chen2026dflash} as a \textit{unigram proposer} to independently generate candidate tokens for every position in a drafting block. Second, a redesigned \textit{bigram selector} scores adjacent candidate-token pairs conditioned on the proposer hidden states. It derives predecessor and successor representations from frozen target embeddings through separate MLPs and learns depth-dependent weights for combining proposer logits and selector scores. Third, we use the resulting transition scores to construct the verification tree. By simplifying the token-wise dependency into bigram correlation, we overcome the challenge of parallelizing token-wise dependency in tree drafting.
    \item \textbf{Training.} Unlike other drafters trained on target-regenerated rollouts at the inference temperature, we use a renormalized KL training objective on high-temperature regenerated target trajectories to diversify the positive training labels. This \textit{tree-native} training strategy allocates more supervision signals to plausible alternative paths in the target distribution and distinguishes them better from low-quality alternative paths in tree construction, thereby relieving the train--inference mismatch in tree drafting.
\end{itemize}
We extensively evaluate UBTree on both academic and production-scale benchmarks. UBTree outperforms all existing methods on both academic-scale and production benchmarks. It achieves an average decoding speedup of $5.84$--$6.94\times$ over autoregressive decoding on Qwen3-4B and Qwen3-8B with increased acceptance lengths. It also rivals frontier baselines such as DSpark~\citep{cheng2026dspark} in the acceleration of production LLMs like Ling3-Flash-124B. Additionally, UBTree shows outstanding durability on long-context tasks compared to baselines. In summary, UBTree provides a frontier solution for accelerating LLMs in both academic settings and production environments.

\section{Background}
  \label{sec:method_prelim}
  \phantomsection
  \noindent\textbf{Speculative Decoding.}
  \label{sec:prelim_spec_decoding}
  Given a verified prefix $x_{\le t}$, a lightweight draft model proposes $\gamma$ future tokens, which the target model verifies in one forward pass~\citep{leviathan2023fast,chen2023accelerating}.
  Each round accepts consecutive draft tokens from the start of the proposed block until the first rejection.
  Let $\tau\in[1,\gamma+1]$ denote the average of accepted tokens per round or \textit{acceptance length}, including a token $x_0$ prefilled by the last verification.
  With $T_{\mathrm{draft}}$ covering proposal construction and $T_{\mathrm{verify}}$ covering verification and commit work, the average per-token \textit{latency} and \textit{speedup} are~\citep{chen2026dflash,huang2026domino}
  \begin{equation}
      L_{\mathrm{spec}}=\frac{T_{\mathrm{draft}}+T_{\mathrm{verify}}}{\tau},
      \qquad
      \eta=\frac{L_{\mathrm{target}}}{L_{\mathrm{spec}}},
      \label{eq:spec-latency}
  \end{equation}
  where $L_{\mathrm{target}}$ is the per-token latency of target-only autoregressive decoding.
  \phantomsection
  \noindent\textbf{DFlash \& DFlash 2.}
  \label{sec:prelim-dflash2}
  DFlash~\citep{chen2026dflash} proposes a $\gamma$-token block in parallel, conditioned on target hidden features through key-value context of one target prefilling forward. We denote the prefilled token $x_0$ by \textit{anchor token}. For each future draft position $i=1,...,\gamma$, one proposer forward pass produces final hidden state $h_i$, which is subsequently mapped to proposer logit $u_i$ through the frozen target LM head. The draft tokens of DFlash are predicted in parallel from masks where little token-wise dependencies are involved. DFlash~2~\citep{inco2026dflash2} fixes this by adding convolutions to the DFlash backbone and a low-rank bigram selector. For predecessor $a$ and successor $b$ at position $i$, the selector score $\delta_i(a,b)$ is added to the proposer logit $u_i(b)$ to obtain a combined score:
  \begin{equation}
      \delta_i(a,b)=\bigl(P(h_i)\odot \phi(a)\bigr)^\top \psi(b),
      \qquad
      u_i'(a,b)=u_i(b)+\delta_i(a,b),
      \label{eq:dflash2-selector}
  \end{equation}
  where $\odot$ stands for Hadamard product. Here, the linear projector $P$ maps $h_i$ to a low-dimensional space, and $\phi$ and $\psi$ are low-rank embeddings for predecessor and successor tokens, respectively. The selector scores all adjacent pairs from the position-wise top-$K$ candidate sets in parallel; token selection then follows a single chain using the combined scores.

  \phantomsection
  \noindent\textbf{Tree Drafting.}
  \label{sec:prelim-tree}
  DFlash and DFlash 2 are \textit{single-path} drafting methods, which propose a single $\gamma$-token draft path to match a $\gamma$-token target path for the target model to verify. \textit{Tree drafting}, by contrast, constructs a draft tree storing multiple candidate draft paths. Taking candidate tokens as nodes, different paths share nodes across continuations with common prefixes. We then use tree attention to verify the whole draft tree in one target-model pass and picks the path with the longest accepted token continuations by the target model in the tree~\citep{miao2024specinfer,cai2024medusa}. This verification procedure maintains the lossless property of speculative decoding, and accepted path need not be the proposer's top-ranked.

  DDTree~\citep{ringel2026accelerating} and DARTree~\citep{li2026dartree} construct budgeted trees from block-parallel proposals. DDTree uses best-first search, ranking paths by accumulated position-wise token log-probabilities. DARTree corrects token log-probabilities by a lightweight autoregressive head and ranks the cumulative log-probabilities with a depth decay and a $W$ node budget at each depth. These retained nodes accumulate into a candidate supertree of maximum depth $\gamma$, from which the $B$ highest-scoring non-root nodes are selected for verification. Correction at the next depth still depends on the tokens selected at the preceding depth, leading to extra overheads. 
\begin{figure}[!t]
    \centering
    \includegraphics[width=\linewidth]{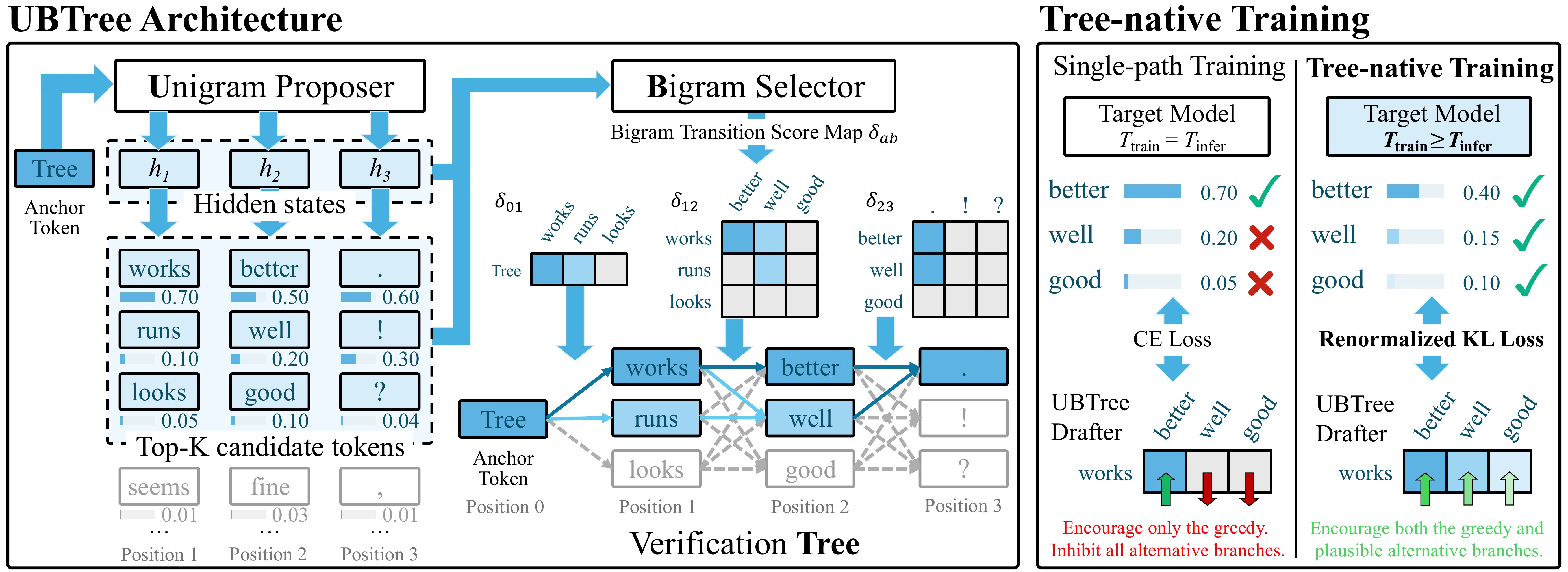}
    \caption{Overview of UBTree. \textbf{Left: UBTree Architecture.} The unigram proposer generates top-$K$ candidates at each draft position in parallel. The bigram selector scores adjacent candidate transitions in parallel, and tree construction uses these precomputed scores to build a verification tree. \textbf{Right: Tree-native Training.} Our training combines high-temperature target traces with renormalized KL supervision to encourage plausible alternative branches.}
    \label{fig:ubtree-overview}
\end{figure}

\section{Method}
\label{sec:method}

Two key designs of UBTree are 1) parallelizing token-wise dependency modeling in tree drafting and 2) training the tree drafter towards optimal tree drafting. Section~\ref{sec:method-architecture} first explains how we incorporate token-dependencies with fully parallelized neural compuation in tree drafting. We use two parallel neural forwards to propose candidate tokens at each block position (Section~\ref{sec:method-unigram}) and select transition paths among these candidates (Section~\ref{sec:method-selector}). The verification tree is then built from paths selected by the selector (Section~\ref{sec:method-tree}). Section~\ref{sec:method-training} presents our \textit{tree-native} training, which regenerates the target rollouts for training under high temperatures and trains the drafter with a renormalized KL objective. We summarize these two designs of UBTree in Figure~\ref{fig:ubtree-overview}.

\subsection{UBTree Architecture}
\label{sec:method-architecture}
\subsubsection{Unigram Proposer}
\label{sec:method-unigram}
The unigram proposer is designed to propose candidate tokens at each position without considering cross-token dependencies. Let $c$ be the key-value context from verified prefixes and $x_0$ be the prefilled anchor token. The unigram proposer can be formalized by
\begin{equation}
    q_i(x_{i}\mid x_0,c)=\operatorname{softmax}(u_i(x
    _0,c)),\quad i=1,2,...,\gamma,
    \label{eq:ubtree-unigram}
\end{equation}
where $i$ denotes the position in the block. Since the proposal at each position does not depend on each other, its neural forward could be easily parallelized. In practice, we deploy a pretrained DFlash backbone~\citep{chen2026dflash} as our unigram proposer, where its output logits are $u_i(x_0,c)$ and probabilities are $q_i(x_{i}\mid x_0,c)$. We also denote its predicted hidden states by $h_i$. We do not use DFlash2 backbone~\citep{inco2026dflash2} because it differs from DFlash backbone by strengthening token-wise connection, which is not the task of our unigram proposal.

\noindent\textbf{Candidate Sets.}
We pick the tokens with top-$K$ probabilities in $q_i$ at each position as candidate tokens of this position $\mathcal C_i$, with $\mathcal C_0=\{x_t\}$. The Cartesian product $\mathcal C_1\times\cdots\times\mathcal C_\gamma$ forms $K^\gamma$ candidate paths. Enumerating these paths is infeasible, so UBTree scores their adjacent transitions and searches a bounded subset by the bigram selector in the next section.

\subsubsection{Bigram Selector}
\label{sec:method-selector}
Inspired by DFlash~2~\citep{inco2026dflash2}, we incorporate a lightweight bigram selector scores adjacent predecessor--successor token pairs from candidates, conditioned on the proposer hidden state $h_i$ at the successor position. This trainable selector models token-wise dependencies in a parallel manner. To improve the selector, we redesign its architecture with deep codebooks and depth calibration.

\noindent\textbf{Deep Codebooks.}
For predecessor token $a$ and successor candidate $b$ at position $i$, DFlash~2 learns two vocabulary-wide codebooks $\phi$ and $\psi$ by two embedding layers (See~\eqref{eq:dflash2-selector}). UBTree instead derives the codebooks from the frozen target token embedding $e_v\in\mathbb R^{d_e}$ and two bias-free MLPs:
\begin{equation}
  \phi(v)=W_{\phi,2}\,\operatorname{SiLU}(W_{\phi,1}e_v),
  \qquad
  \psi(v)=W_{\psi,2}\,\operatorname{SiLU}(W_{\psi,1}e_v).
  \label{eq:ubtree-codebook}
\end{equation}
At inference, we precompute both MLPs once and store them as a vocabulary-wide lookup table. While keeping inference efficiency, our deep codebook introduces nonlinearity and makes use of features in the target embedding. Hence, it improves the selector performance.

\noindent\textbf{Depth Calibration.}
DFlash~2 combines proposer logits and selector scores with fixed unit weights. However, their reliability can vary across draft depths. UBTree therefore learns two positive, depth-dependent scales:
\begin{equation}
    s_i(a,b)=\alpha_i u_i(b)+\lambda_i\delta_i(a,b),
    \qquad
    \alpha_i=\exp(\rho_i),\quad
    \lambda_i=\exp(\kappa_i).
    \label{eq:ubtree-depth}
\end{equation}
The ratio $\lambda_i/\alpha_i$ controls the weight of selector scores relative to proposer logits, while their common magnitude controls the concentration of the normalized successor distribution. These scales depend only on depth and preserve parallel edge scoring.

\noindent\textbf{Parallel Computation.}
Once the proposer provides the candidate sets and hidden states, we compute transition scores for all $K+(\gamma-1)K^2$ predecessor-successor token pairs in parallel. At depth $i$, let $\boldsymbol{\Phi}_{i-1}$ have rows $\phi(a)^\top$ for predecessor candidates $a\in\mathcal C_{i-1}$, and let $\boldsymbol{\Psi}_i$ have rows $\psi(b)^\top$ for successor candidates $b\in\mathcal C_i$. With proposer's hidden state $h_i$ at the successor position, we obtain
\begin{equation}
    \boldsymbol\Delta_i
    =\boldsymbol{\Phi}_{i-1}\operatorname{diag}\!\bigl(P(h_i)\bigr)\boldsymbol{\Psi}_i^\top
    \in\mathbb R^{|\mathcal C_{i-1}|\times K}.
    \label{eq:ubtree-edge-matrix}
\end{equation}
Here, $\operatorname{diag}(\cdot)$ places the input vector on the diagonal. The entry at the row for $a$ and column for $b$ is the selector score $\delta_i(a,b)$. After combining $\delta_i(a,b)$ with proposer logits $u_i(b)$ via~\eqref{eq:ubtree-depth}, tree construction can proceed depth by depth sequentially without any neural computation.
\subsubsection{Tree Construction and Verification}
\label{sec:method-tree}

With unique scores for every transition $s_i(a,b)$ in $C_{i-1}\times C_i$, we are able to score all $K^\gamma$ candidate paths consisting of these transitions without neural computation. Specifically, we can normalize $s_i(a,b)$ into log-probabilities over successor candidates, then rank paths by cumulative log-probabilities. However, to save the computation cost, we use DARTree's depth-wise progressive tree construction and global pruning procedure~\citep{li2026dartree} instead of traversing all possible paths. After tree construction, target model will verify the whole tree in one-forward pass via tree attention, and keep the longest accepted continuation for lossless acceleration. Appendix~\ref{app:tree-construction} specifies the normalization, path scoring, search and verification procedure.

\subsection{Tree-native Training}
\label{sec:method-training}
Tree drafting recommends multiple paths for target verification, rather than only proposing the greedy path. Hence, we need to diversify positive training signals to help the selector better recognize alternative plausible paths other than the greedy one. To this end, we use higher temperature $T_{\mathrm{train}}$ to regenerate the training data than the inference temperature $T_{\mathrm{infer}}$ and supervise the training by a renormalized KL objective.

\noindent\textbf{Training with $T_{\mathrm{train}}\ge T_{\mathrm{infer}}$ for different $T_{\mathrm{infer}}$.}
The training temperature $T_{\mathrm{train}}$ controls both the continuation distribution seen by the proposer and selector and the target distribution used for KL supervision. Temperature-matched training would set $T_{\mathrm{train}}=T_{\mathrm{infer}}$. As mentioned above, the selector needs to improve the plausibility of multiple paths in the tree, rather than focusing only on the top path. We can therefore make $T_{\mathrm{train}}\ge T_{\mathrm{infer}}$ to make those high-probability alternative paths more distinguishable. $T_{\mathrm{train}}$ then becomes an independent hyper-parameter to be tuned empirically. The best $T_{\mathrm{train}}$ amplifies the probability gap between plausible and other choices. Interestingly, $T_{\mathrm{train}}$ is disentangled with $T_{\mathrm{infer}}$ to some extent, that one UBTree drafter trained on $T_{\mathrm{train}}$ can be used in different $T_{\mathrm{infer}}$, which simplifies the drafter adaptation. \textbf{We therefore use drafters trained at $T_{\mathrm{train}}=1$ for both $T_{\mathrm{infer}}=0$ and $T_{\mathrm{infer}}=1$ in our experiments.}

\noindent\textbf{Renormalized KL Loss.} To train the selector to better discriminate between alternates other than the top choice, we use the forward KL divergence between the renormalized probabilities~\citep{zhu2026many} of the target and that of the selector over the proposer's top-$K$ candidate tokens instead of the classical cross-entropy (CE) loss. For each valid future position, let $z_i(v)$ be the target logit and $s_i(y_{i-1}^{*},v)$ be the selector logit given the predecessor token $y_{i-1}^{*}$. For $T_{\mathrm{train}}>0$, we define the renormalized target probabilities $\widetilde p_i(v)$ at training temperature $T_{\mathrm{train}}$ and the selector probabilities $\widetilde q_i(v)$ on the top-$K$ token support $\mathcal C_i$:
\begin{equation}
\widetilde p_i(v)
=\frac{\exp(z_i(v)/T_{\mathrm{train}})}
{\sum_{b\in\mathcal C_i}\exp(z_i(b)/T_{\mathrm{train}})},
\qquad
\widetilde q_i(v)
=\frac{\exp(s_i(y_{i-1}^{*},v))}
{\sum_{b\in\mathcal C_i}\exp(s_i(y_{i-1}^{*},b))}.
\label{eq:ubtree-training-distributions}
\end{equation}
For $T_{\mathrm{train}}=0$, the target distribution is understood in the limit $T_{\mathrm{train}}\to0^+$.
The forward KL divergence between $\widetilde p_i(v)$ and $\widetilde q_i(v)$ constitutes the main term of our training objective. An auxiliary cross-entropy term, also used in the first-stage proposer training, is applied solely to the proposer $q_i$ to maintain the candidate distribution. Using weight $\beta$ to balance two terms, the joint objective is
\begin{equation}
\mathcal L
=
\mathrm{KL}(\widetilde p_i|\widetilde q_i)
-\beta\log q_i(y_i^{*}\mid x_{\le t}),
\qquad \beta=0.1.
\label{eq:ubtree-training-loss}
\end{equation}

\section{Experiments}
\label{sec:experiments}

\newcommand{\UBTreeFill}[1]{\textit{[TO FILL: #1]}}

\newcommand{\UBTreeMainHeader}{
\toprule
\multirow{3}{*}{Model} & \multirow{3}{*}{Method}
& \multicolumn{6}{c}{\textsc{Math}}
& \multicolumn{6}{c}{\textsc{Code}}
& \multicolumn{2}{c}{\textsc{Chat}}
& \multicolumn{2}{c}{} \\
\cmidrule(lr){3-8}\cmidrule(lr){9-14}\cmidrule(lr){15-16}
& & \multicolumn{2}{c}{GSM8K}
& \multicolumn{2}{c}{M-500}
& \multicolumn{2}{c}{AIME}
& \multicolumn{2}{c}{HE}
& \multicolumn{2}{c}{MBPP}
& \multicolumn{2}{c}{LCB}
& \multicolumn{2}{c}{MT-B}
& \multicolumn{2}{c}{Avg.} \\
& & {\scriptsize Speedup} & {\scriptsize $\tau$}
& {\scriptsize Speedup} & {\scriptsize $\tau$}
& {\scriptsize Speedup} & {\scriptsize $\tau$}
& {\scriptsize Speedup} & {\scriptsize $\tau$}
& {\scriptsize Speedup} & {\scriptsize $\tau$}
& {\scriptsize Speedup} & {\scriptsize $\tau$}
& {\scriptsize Speedup} & {\scriptsize $\tau$}
& {\scriptsize Speedup} & {\scriptsize $\tau$} \\
\midrule
}

\newcommand{\UBTreeMainRow}[2]{#1 & #2 & &  & &  & &  & &  & &  & &  & &  & &  \\}

\newcommand{\UBTreeMainGroupRow}[2]{#1 & \multicolumn{17}{l}{\textit{#2}} \\}

\newcommand{\UBTreeWideResultHeader}{
\toprule
\multirow{3}{*}{Model} & \multirow{3}{*}{Method}
& \multicolumn{6}{c}{\textsc{Math}}
& \multicolumn{6}{c}{\textsc{Code}}
& \multicolumn{2}{c}{\textsc{Chat}}
& \multicolumn{2}{c}{} \\
\cmidrule(lr){3-8}\cmidrule(lr){9-14}\cmidrule(lr){15-16}
& & \multicolumn{2}{c}{GSM8K}
& \multicolumn{2}{c}{M-500}
& \multicolumn{2}{c}{AIME}
& \multicolumn{2}{c}{HE}
& \multicolumn{2}{c}{MBPP}
& \multicolumn{2}{c}{LCB}
& \multicolumn{2}{c}{MT-B}
& \multicolumn{2}{c}{Avg.} \\
& & {\scriptsize Speedup} & {\scriptsize $\tau$}
& {\scriptsize Speedup} & {\scriptsize $\tau$}
& {\scriptsize Speedup} & {\scriptsize $\tau$}
& {\scriptsize Speedup} & {\scriptsize $\tau$}
& {\scriptsize Speedup} & {\scriptsize $\tau$}
& {\scriptsize Speedup} & {\scriptsize $\tau$}
& {\scriptsize Speedup} & {\scriptsize $\tau$}
& {\scriptsize Speedup} & {\scriptsize $\tau$} \\
\midrule
}

\newcommand{\UBTreeWideResultRow}[2]{#1 & #2 & &  & &  & &  & &  & &  & &  & &  & &  \\}

\newcommand{\UBTreeAblationHeader}{
\toprule
\multirow{3}{*}{}
& \multicolumn{6}{c}{\textsc{Math}}
& \multicolumn{6}{c}{\textsc{Code}}
& \multicolumn{2}{c}{\textsc{Chat}}
& \multicolumn{2}{c}{} \\
\cmidrule(lr){2-7}\cmidrule(lr){8-13}\cmidrule(lr){14-15}
& \multicolumn{2}{c}{GSM8K}
& \multicolumn{2}{c}{M-500}
& \multicolumn{2}{c}{AIME}
& \multicolumn{2}{c}{HE}
& \multicolumn{2}{c}{MBPP}
& \multicolumn{2}{c}{LCB}
& \multicolumn{2}{c}{MT-B}
& \multicolumn{2}{c}{Avg.} \\
& {\scriptsize Speedup} & {\scriptsize $\tau$}
& {\scriptsize Speedup} & {\scriptsize $\tau$}
& {\scriptsize Speedup} & {\scriptsize $\tau$}
& {\scriptsize Speedup} & {\scriptsize $\tau$}
& {\scriptsize Speedup} & {\scriptsize $\tau$}
& {\scriptsize Speedup} & {\scriptsize $\tau$}
& {\scriptsize Speedup} & {\scriptsize $\tau$}
& {\scriptsize Speedup} & {\scriptsize $\tau$} \\
\midrule
}

\newcommand{\UBTreeAblationRow}[1]{#1 & &  & &  & &  & &  & &  & &  & &  & &  \\}

\newcommand{\UBTreeTemperatureHeader}{
\toprule
\multirow{3}{*}{$T_{\mathrm{train}}$} & \multirow{3}{*}{Variant}
& \multicolumn{6}{c}{\textsc{Math}}
& \multicolumn{6}{c}{\textsc{Code}}
& \multicolumn{2}{c}{\textsc{Chat}}
& \multicolumn{2}{c}{} \\
\cmidrule(lr){3-8}\cmidrule(lr){9-14}\cmidrule(lr){15-16}
& & \multicolumn{2}{c}{GSM8K}
& \multicolumn{2}{c}{M-500}
& \multicolumn{2}{c}{AIME}
& \multicolumn{2}{c}{HE}
& \multicolumn{2}{c}{MBPP}
& \multicolumn{2}{c}{LCB}
& \multicolumn{2}{c}{MT-B}
& \multicolumn{2}{c}{Avg.} \\
& & {\scriptsize Speedup} & {\scriptsize $\tau$}
& {\scriptsize Speedup} & {\scriptsize $\tau$}
& {\scriptsize Speedup} & {\scriptsize $\tau$}
& {\scriptsize Speedup} & {\scriptsize $\tau$}
& {\scriptsize Speedup} & {\scriptsize $\tau$}
& {\scriptsize Speedup} & {\scriptsize $\tau$}
& {\scriptsize Speedup} & {\scriptsize $\tau$}
& {\scriptsize Speedup} & {\scriptsize $\tau$} \\
\midrule
}

\newcommand{\UBTreeTemperatureRow}[2]{#1 & #2 & &  & &  & &  & &  & &  & &  & &  & &  \\}

We evaluate UBTree on benchmarks and report decoding speedup $\eta$ over autoregressive baseline and average acceptance length $\tau$. Avg. means average results across all listed benchmarks.
\begin{table*}[t]
\centering
\caption{Speedup and average acceptance length ($\tau$) on instruct and reasoning models. All $\tau$ values include the target-produced bonus token. M-500, HE, LCB, and MT-B denote MATH-500, HumanEval, LiveCodeBench, and MT-Bench. $\dagger$ values are paper-reported.}
\label{tab:main-results}
\begingroup
\small
\setlength{\tabcolsep}{2pt}
\renewcommand{\arraystretch}{0.95}
\resizebox{\textwidth}{!}{
\begin{tabular}{@{}c l @{\hspace{1.0em}} cc cc cc @{\hspace{1.0em}} cc cc cc @{\hspace{1.0em}} cc @{\hspace{1.0em}} cc@{}}
\toprule
\multicolumn{18}{@{}l}{\textbf{Instruct Models}} \\
\UBTreeMainHeader
\UBTreeMainGroupRow{\multirow{9}{*}{\shortstack{Qwen3-4B\\$T_{\mathrm{infer}}=0$}}}{Single-path}
 & DFlash (16) & 5.95 & 8.57 & 5.76 & 8.34 & 4.45 & 6.34 & 4.28 & 6.08 & 4.03 & 5.82 & 3.83 & 5.39 & 2.69 & 4.60 & 4.43 & 6.45 \\
 & Domino\textsuperscript{\ensuremath{\dagger}} (17) & 8.02 & 10.12 & 7.14 & 9.01 & 5.97 & 7.49 & 5.66 & 7.02 & 5.59 & 7.11 & 5.33 & 6.86 & 3.28 & 5.12 & 5.86 & 7.53 \\
\UBTreeMainGroupRow{}{Tree-based}
 & EAGLE-3 (16) & 4.04 & 7.20 & 3.97 & 6.96 & 3.45 & 6.05 & 3.46 & 6.07 & 3.42 & 6.05 & 3.19 & 5.53 & 2.40 & 4.67 & 3.42 & 6.08 \\
 & EAGLE-3 (64) & 4.42 & 7.97 & 4.45 & 7.85 & 4.02 & 7.11 & 4.00 & 7.07 & 3.95 & 7.11 & 3.75 & 6.58 & 2.80 & 5.46 & 3.91 & 7.02 \\
 & DDTree (64) & 6.69 & 10.22 & 6.62 & 10.10 & 5.51 & 8.49 & 5.18 & 7.89 & 5.10 & 7.80 & 4.78 & 7.23 & 3.51 & 6.14 & 5.34 & 8.27 \\
 & DARTree (64) & 7.89 & 12.50 & 7.63 & 11.90 & 6.55 & \textbf{10.37} & 6.09 & 9.78 & 6.33 & \textbf{10.07} & 5.75 & 9.04 & 3.93 & 7.23 & 6.31 & 10.13 \\
 & \textbf{UBTree (64)} & \textbf{8.54} & \textbf{12.58} & \textbf{8.18} & \textbf{12.15} & \textbf{6.92} & 10.13 & \textbf{7.07} & \textbf{10.22} & \textbf{6.66} & 9.98 & \textbf{6.37} & \textbf{9.24} & \textbf{4.39} & \textbf{7.32} & \textbf{6.88} & \textbf{10.23} \\
\midrule
\UBTreeMainGroupRow{\multirow{9}{*}{\shortstack{Qwen3-4B\\$T_{\mathrm{infer}}=1$}}}{Single-path}
 & DFlash (16) & 5.28 & 7.67 & 4.56 & 6.89 & 3.10 & 4.64 & 4.09 & 5.86 & 3.84 & 5.52 & 3.25 & 4.62 & 2.52 & 4.26 & 3.81 & 5.64 \\
 & Domino\textsuperscript{\ensuremath{\dagger}} (17) & 6.79 & 8.61 & 5.67 & 7.35 & 3.75 & 4.83 & 5.02 & 6.30 & 4.96 & 6.36 & 5.04 & 6.45 & 3.02 & 4.56 & 4.89 & 6.35 \\
\UBTreeMainGroupRow{}{Tree-based}
 & EAGLE-3 (16) & 3.66 & 6.61 & 3.51 & 6.36 & 2.72 & 4.88 & 3.34 & 5.93 & 3.21 & 5.80 & 2.80 & 4.94 & 2.21 & 4.34 & 3.06 & 5.55 \\
 & EAGLE-3 (64) & 4.13 & 7.66 & 3.96 & 7.29 & 3.20 & 6.00 & 3.75 & 6.84 & 3.71 & 6.92 & 3.27 & 5.96 & 2.54 & 5.12 & 3.51 & 6.54 \\
 & DDTree (64) & 6.19 & 9.49 & 5.62 & 8.76 & 4.14 & 6.31 & 5.17 & 7.81 & 4.90 & 7.50 & 4.19 & 6.37 & 3.37 & 5.66 & 4.80 & 7.41 \\
 & DARTree (64) & 6.94 & 11.01 & 6.05 & 9.73 & 4.10 & 6.99 & 5.84 & 9.10 & 5.76 & 9.28 & 4.71 & 7.33 & 3.64 & 6.53 & 5.29 & 8.57 \\
 & \textbf{UBTree (64)} & \textbf{7.95} & \textbf{11.79} & \textbf{7.13} & \textbf{10.54} & \textbf{5.10} & \textbf{7.63} & \textbf{6.42} & \textbf{9.43} & \textbf{6.15} & \textbf{9.30} & \textbf{5.27} & \textbf{7.72} & \textbf{4.02} & \textbf{6.68} & \textbf{6.00} & \textbf{9.01} \\
\midrule
\UBTreeMainGroupRow{\multirow{9}{*}{\shortstack{Qwen3-8B\\$T_{\mathrm{infer}}=0$}}}{Single-path}
 & DFlash (16) & 5.98 & 8.59 & 5.84 & 8.51 & 4.75 & 6.73 & 4.46 & 6.31 & 4.07 & 5.87 & 3.91 & 5.48 & 2.64 & 4.54 & 4.52 & 6.58 \\
 & Domino\textsuperscript{\ensuremath{\dagger}} (17) & 7.92 & 10.03 & 7.38 & 9.43 & 5.85 & 7.41 & 5.89 & 7.39 & 5.53 & 7.04 & 5.27 & 7.04 & 3.29 & 5.18 & 5.88 & 7.65 \\
\UBTreeMainGroupRow{}{Tree-based}
 & EAGLE-3 (16) & 4.23 & 7.43 & 4.20 & 7.30 & 3.71 & 6.46 & 3.68 & 6.42 & 3.62 & 6.39 & 3.54 & 6.07 & 2.53 & 4.88 & 3.64 & 6.42 \\
 & EAGLE-3 (64) & 4.57 & 8.20 & 4.66 & 8.11 & 4.34 & 7.51 & 4.23 & 7.39 & 4.13 & 7.39 & 4.11 & 7.10 & 2.99 & 5.77 & 4.15 & 7.35 \\
 & DDTree (64) & 6.65 & 10.19 & 6.60 & 10.12 & 5.64 & 8.63 & 5.43 & 8.17 & 5.07 & 7.68 & 4.92 & 7.30 & 3.49 & 6.09 & 5.40 & 8.31 \\
 & DARTree (64) & 8.26 & 12.49 & 7.61 & 12.13 & 6.71 & \textbf{10.37} & 6.41 & 10.09 & 6.32 & \textbf{10.02} & 5.88 & 9.19 & 4.01 & \textbf{7.33} & 6.46 & \textbf{10.23} \\
 & \textbf{UBTree (64)} & \textbf{8.65} & \textbf{12.58} & \textbf{8.46} & \textbf{12.30} & \textbf{7.05} & 10.17 & \textbf{7.12} & \textbf{10.12} & \textbf{6.64} & 9.79 & \textbf{6.43} & \textbf{9.24} & \textbf{4.25} & 7.25 & \textbf{6.94} & 10.21 \\
\midrule
\UBTreeMainGroupRow{\multirow{9}{*}{\shortstack{Qwen3-8B\\$T_{\mathrm{infer}}=1$}}}{Single-path}
 & DFlash (16) & 5.05 & 7.37 & 4.57 & 6.77 & 2.99 & 4.47 & 3.74 & 5.26 & 3.64 & 5.20 & 3.14 & 4.41 & 2.34 & 3.96 & 3.64 & 5.35 \\
 & Domino\textsuperscript{\ensuremath{\dagger}} (17) & 6.47 & 8.34 & 5.40 & 7.20 & 3.78 & 4.92 & 4.75 & 5.98 & 4.73 & 6.02 & 5.06 & 6.72 & 2.94 & 4.61 & 4.73 & 6.26 \\
\UBTreeMainGroupRow{}{Tree-based}
 & EAGLE-3 (16) & 3.93 & 7.07 & 3.74 & 6.68 & 3.01 & 5.30 & 3.35 & 5.89 & 3.35 & 6.01 & 3.01 & 5.25 & 2.36 & 4.55 & 3.25 & 5.82 \\
 & EAGLE-3 (64) & 4.39 & 8.05 & 4.23 & 7.65 & 3.51 & 6.34 & 3.88 & 7.00 & 3.88 & 7.14 & 3.53 & 6.32 & 2.73 & 5.45 & 3.73 & 6.85 \\
 & DDTree (64) & 6.10 & 9.23 & 5.61 & 8.68 & 4.04 & 6.31 & 4.91 & 7.37 & 4.63 & 7.02 & 4.14 & 6.22 & 3.23 & 5.48 & 4.67 & 7.19 \\
 & DARTree (64) & 7.04 & 11.01 & 5.86 & 9.59 & 4.48 & 7.21 & 5.26 & 8.36 & 5.53 & \textbf{8.91} & 4.45 & 7.06 & 3.59 & \textbf{6.48} & 5.17 & 8.37 \\
 & \textbf{UBTree (64)} & \textbf{7.97} & \textbf{11.55} & \textbf{6.99} & \textbf{10.44} & \textbf{4.99} & \textbf{7.43} & \textbf{6.17} & \textbf{8.85} & \textbf{5.92} & 8.80 & \textbf{5.10} & \textbf{7.47} & \textbf{3.77} & 6.36 & \textbf{5.84} & \textbf{8.70} \\
\bottomrule
\end{tabular}
}
\vspace{2pt}

\resizebox{\textwidth}{!}{
\begin{tabular}{@{}c l @{\hspace{1.0em}} cc cc cc @{\hspace{1.0em}} cc cc cc @{\hspace{1.0em}} cc @{\hspace{1.0em}} cc@{}}
\toprule
\multicolumn{18}{@{}l}{\textbf{Reasoning Models}} \\
\UBTreeMainHeader
\UBTreeMainGroupRow{\multirow{7}{*}{\shortstack{Qwen3-4B\\$T_{\mathrm{infer}}=0$}}}{Single-path}
 & DFlash (16) & 5.18 & 7.28 & 4.53 & 6.45 & 3.67 & 5.26 & 3.91 & 5.46 & 3.72 & 5.17 & 3.37 & 4.68 & 2.99 & 4.64 & 3.91 & 5.56 \\
\UBTreeMainGroupRow{}{Tree-based}
 & EAGLE-3 (16) & 3.49 & 6.01 & 3.04 & 5.23 & 2.65 & 4.62 & 2.82 & 4.83 & 2.80 & 4.78 & 2.43 & 4.16 & 2.25 & 4.11 & 2.78 & 4.82 \\
 & EAGLE-3 (64) & 4.19 & 7.24 & 3.75 & 6.48 & 3.29 & 5.77 & 3.52 & 6.05 & 3.51 & 6.04 & 3.04 & 5.28 & 2.74 & 5.07 & 3.44 & 5.99 \\
 & DDTree (64) & 6.09 & 9.19 & 5.54 & 8.42 & 4.77 & 7.20 & 4.85 & 7.29 & 4.68 & 7.01 & 4.31 & 6.45 & 3.82 & 6.15 & 4.87 & 7.39 \\
 & \textbf{UBTree (64)} & \textbf{7.64} & \textbf{10.95} & \textbf{6.99} & \textbf{10.01} & \textbf{5.73} & \textbf{8.53} & \textbf{6.21} & \textbf{8.85} & \textbf{5.98} & \textbf{8.54} & \textbf{5.45} & \textbf{7.75} & \textbf{4.71} & \textbf{7.22} & \textbf{6.10} & \textbf{8.84} \\
\midrule
\UBTreeMainGroupRow{\multirow{7}{*}{\shortstack{Qwen3-4B\\$T_{\mathrm{infer}}=1$}}}{Single-path}
 & DFlash (16) & 4.80 & 6.77 & 4.36 & 6.18 & 3.57 & 5.15 & 3.76 & 5.23 & 3.55 & 4.93 & 3.21 & 4.47 & 2.89 & 4.40 & 3.73 & 5.30 \\
\UBTreeMainGroupRow{}{Tree-based}
 & EAGLE-3 (16) & 3.01 & 5.28 & 2.78 & 4.86 & 2.41 & 4.22 & 2.54 & 4.42 & 2.47 & 4.28 & 2.24 & 3.92 & 2.00 & 3.69 & 2.49 & 4.38 \\
 & EAGLE-3 (64) & 3.64 & 6.57 & 3.34 & 6.00 & 2.98 & 5.37 & 3.17 & 5.66 & 3.06 & 5.48 & 2.77 & 4.99 & 2.35 & 4.49 & 3.04 & 5.51 \\
 & DDTree (64) & 4.87 & 7.35 & 4.69 & 7.11 & 4.01 & 6.10 & 4.17 & 6.22 & 3.97 & 5.93 & 3.57 & 5.37 & 3.28 & 5.11 & 4.08 & 6.17 \\
 & \textbf{UBTree (64)} & \textbf{6.56} & \textbf{9.50} & \textbf{6.12} & \textbf{8.78} & \textbf{5.12} & \textbf{7.47} & \textbf{5.44} & \textbf{7.74} & \textbf{5.18} & \textbf{7.41} & \textbf{4.57} & \textbf{6.61} & \textbf{4.10} & \textbf{6.17} & \textbf{5.30} & \textbf{7.67} \\
\bottomrule
\end{tabular}
}
\endgroup
\vspace{0cm}
\end{table*}

\subsection{Academic-scale Experiment}
\label{sec:exp-main}
\label{sec:exp-setup}

\noindent\textbf{Benchmarks.}
We evaluate UBTree on Qwen3-4B and Qwen3-8B~\citep{yang2025qwen3} with thinking disabled (instruct models), and on Qwen3-4B with thinking enabled (reasoning models), under $T_{\mathrm{infer}}=\{0,1\}$. The benchmarks cover math (GSM8K~\citep{cobbe2021training}, MATH-500~\citep{lightman2024let}, and AIME), code (HumanEval~\citep{chen2021evaluating}, MBPP~\citep{austin2021program}, and LiveCodeBench~\citep{jain2025livecodebench}), and dialogue (MT-Bench~\citep{zheng2023judging}).

\noindent\textbf{Implementation.}
We train UBTree on OpenPerfectBlend~\citep{xu2024perfect} with regenerated target responses under $T_{\mathrm{train}}=1$. We retrain DFlash~\citep{chen2026dflash} proposer with block size 16 with SpecForge~\citep{li2026specforge} for 6 epochs till convergence, and then jointly train the proposer and selector for 2 epochs, with selector and backbone learning rates of $6\times10^{-4}$ and $2\times10^{-5}$, respectively. Unlike baselines that need different checkpoints for different decoding temperatures $T_{\mathrm{infer}}$, we use the same checkpoint of UBTree trained under $T_{\mathrm{train}}=1$ is used at $T_{\mathrm{infer}}=\{0,1\}$. Our evaluations use $1\times$NVIDIA Hopper GPU with batch size 1 and the \textit{transformers} backend. For tree construction, we follow DARTree~\citep{li2026dartree} with $B=64$ non-root nodes, a search width of $W=12$, and $K=64$ candidate tokens per position, but use a different depth bonus of $\zeta=0$. Training and evaluation protocols are detailed in Appendix~\ref{app:experimental-setup}.

\noindent\textbf{Baselines.}
We compare UBTree with single-path methods DFlash~\citep{chen2026dflash} and Domino~\citep{huang2026domino}, and tree-based EAGLE-3~\citep{li2026eagle}, DDTree~\citep{ringel2026accelerating}, and DARTree~\citep{li2026dartree}, with budgets in Table~\ref{tab:main-results}. We retrain EAGLE-3 and DFlash on the same OpenPerfectBlend dataset for 6 epochs at corresponding $T_{\mathrm{infer}}$ and report Domino results (with block size 17) from its original paper, trained on OpenPerfectBlend as well. We re-evaluate DDTree and DARTree using retrained DFlash and released Domino checkpoints, respectively, under their original setups. Due to lack of released checkpoints, Domino and DARTree are compared only for instruct models. For fair comparison, we cap the maximum draft depth of DARTree at 16 to match DDTree and UBTree, since the underlying Domino checkpoint uses a block size of 17 and would otherwise permit a larger maximum acceptance length.

\noindent\textbf{Results.}
\label{sec:exp-main-results}
Table~\ref{tab:main-results} shows that UBTree achieves average speedups of $6.88\times$ and $6.94\times$ on Qwen3-4B and Qwen3-8B under greedy decoding, and $6.00\times$ and $5.84\times$ under temperature-1 sampling. It outperforms DARTree over all 28 benchmarks in speedups, improving the average by $7.4$--$13.4\%$. With thinking enabled on Qwen3-4B, UBTree outperforms DDTree by $25.3\%$ and $29.9\%$ in average speedup at $T_{\mathrm{infer}}=0$ and $1$, respectively. Our superiority especially stands out under $T_{\mathrm{infer}}=1$, validating the effectiveness of our high entropy adaptions.

\noindent\textbf{Concurrent Serving.}
\label{sec:exp-main-concurrency}
Table~\ref{tab:concurrency-results} compares UBTree and baselines in SGLang~\citep{zheng2024sglang} at client concurrency $C\in\{1,8,16,32\}$. All methods use SGLang in BF16 on a single GPU with CUDA Graphs. Results are aggregated from the same benchmarks as in Section~\ref{sec:exp-setup}. Following DARTree~\citep{li2026dartree}, we employ an \textit{adaptive-tree} strategy that selects different tree budget $B$ and width $W$ according to the serving load. We tune \(B/W\) separately for each method and concurrency using the same candidate configuration set and selection criterion. At $C\in\{1,8,16,32\}$, UBTree uses $B/W=64/12,64/12,64/12,16/3$ for Qwen3-4B and $B/W=64/12,64/12,16/12,16/3$ for Qwen3-8B and DARTree uses $B/W=64/12,64/12,32/8,16/3$ on both models. UBTree achieves the highest throughput in all settings, outperforming the next-best method by $7.0$--$16.3\%$. At $C=32$, UBTree reaches 6560.4 and 5851.4 tokens/s on Qwen3-4B and Qwen3-8B, respectively, exceeding the strongest single-path baseline by $7.3\%$ and $7.7\%$.
\begin{table*}[t]
\centering
\caption{Concurrent-serving results on Qwen3-4B and Qwen3-8B under $T_{\mathrm{infer}} = 1$. Each cell reports aggregate output tokens/s / speedup over the AR baseline row at the same concurrency.}
\label{tab:concurrency-results}
\begingroup
\scriptsize
\setlength{\tabcolsep}{7pt}
\renewcommand{\arraystretch}{0.95}
\begin{tabular}{@{}clcccc@{}}
\toprule
Model & Method & $C=1$ & $C=8$ & $C=16$ & $C=32$ \\
\midrule
\multirow{6}{*}{Qwen3-4B}
& Autoregressive baseline                         & 253.5 / $1.00\times$  & 1404.9 / $1.00\times$ & 2208.3 / $1.00\times$ & 3307.4 / $1.00\times$ \\
\cmidrule(l){2-6}
& DFlash                      & 796.4 / $3.14\times$   & 2287.2 / $1.63\times$  & 3664.4 / $1.66\times$ & 6071.9 / $1.84\times$ \\
& Domino                      & 816.3 / $3.22\times$   & 2258.9 / $1.61\times$  & 3636.5 / $1.65\times$ & 6115.5 / $1.85\times$ \\
& DDTree                      & 844.6 / $3.33\times$   & 2390.8 / $1.70\times$  & 3669.8 / $1.66\times$ & 4810.4 / $1.45\times$ \\
& DARTree (adaptive $B/W$)   & 978.5 / $3.86\times$   & 2405.1 / $1.71\times$  & 3770.3 / $1.71\times$ & 5619.9 / $1.70\times$ \\
& UBTree (adaptive $B/W$)    & \textbf{1089.1 / $4.30\times$} & \textbf{2797.4 / $1.99\times$} & \textbf{4271.9 / $1.93\times$} & \textbf{6560.4 / $1.98\times$} \\
\midrule
\multirow{6}{*}{Qwen3-8B}
& Autoregressive baseline                          & 180.9 / $1.00\times$   & 1126.2 / $1.00\times$  & 1865.9 / $1.00\times$  & 2770.9 / $1.00\times$ \\
\cmidrule(l){2-6}
& DFlash                      & 634.7 / $3.51\times$   & 2181.7 / $1.94\times$  & 3596.0 / $1.93\times$  & 5265.4 / $1.90\times$ \\
& Domino                      & 676.2 / $3.74\times$   & 2137.9 / $1.90\times$  & 3582.7 / $1.92\times$  & 5435.2 / $1.96\times$ \\
& DDTree                      & 686.8 / $3.80\times$   & 2013.9 / $1.79\times$  & 3303.5 / $1.77\times$  & 3603.6 / $1.30\times$ \\
& DARTree (adaptive $B/W$)   & 824.4 / $4.56\times$   & 2280.1 / $2.02\times$  & 3514.5 / $1.88\times$  & 4844.9 / $1.75\times$ \\
& UBTree (adaptive $B/W$)    & \textbf{881.7 / $4.87\times$} & \textbf{2644.4 / $2.35\times$} & \textbf{4069.9 / $2.18\times$} & \textbf{5851.4 / $2.11\times$} \\
\bottomrule
\end{tabular}
\endgroup
\end{table*}

\begin{table*}[!th]
\centering
\caption{Speedup and acceptance length ($\tau$) on production-scale models. Each benchmark reports speedup and average acceptance length, and the Avg. columns are means over all benchmarks. M-500, HE, LCB, and MT-B denote MATH-500, HumanEval, LiveCodeBench, and MT-Bench.}
\label{tab:production-results}
\begingroup
\small
\setlength{\tabcolsep}{2pt}
\renewcommand{\arraystretch}{0.95}
\resizebox{\textwidth}{!}{
\begin{tabular}{@{}c l @{\hspace{1.0em}} cc cc cc @{\hspace{1.0em}} cc cc cc @{\hspace{1.0em}} cc @{\hspace{1.0em}} cc@{}}
\UBTreeWideResultHeader
\multirow{3}{*}{Ling3-Tiny}
& DSpark & 2.72 & 7.53 & 2.83 & 6.96 & 2.26 & 5.37 & 2.91 & 7.59 & 2.68 & 7.16 & 2.03 & 4.84 & 1.90 & 4.71 & 2.48 & 6.31 \\
& DFlash & 2.66 & 6.57 & 2.64 & 5.74 & 2.26 & 4.76 & 2.77 & 6.39 & 2.47 & 5.87 & 1.95 & 4.11 & 1.86 & 4.11 & 2.37 & 5.36 \\
& \textbf{UBTree} & \textbf{3.05} & \textbf{9.53} & \textbf{3.52} & \textbf{9.86} & \textbf{3.15} & \textbf{8.44} & \textbf{3.41} & \textbf{10.04} & \textbf{3.06} & \textbf{9.35} & \textbf{2.79} & \textbf{7.46} & \textbf{2.30} & \textbf{6.49} & \textbf{3.04} & \textbf{8.74} \\
\midrule
\multirow{3}{*}{Ling3-Flash}
& DSpark & 2.90 & 8.00 & 3.25 & 7.69 & 2.88 & 6.39 & 3.01 & 8.03 & 2.73 & 6.89 & 2.48 & 5.53 & 1.82 & 4.11 & 2.73 & 6.66 \\
& DFlash & 2.93 & 7.44 & 3.28 & 7.07 & 2.86 & 5.75 & 3.13 & 7.79 & 3.01 & 7.01 & 2.54 & 5.13 & 1.83 & 3.73 & 2.80 & 6.27 \\
& \textbf{UBTree} & \textbf{3.23} & \textbf{9.70} & \textbf{3.76} & \textbf{9.58} & \textbf{3.50} & \textbf{8.18} & \textbf{3.48} & \textbf{10.40} & \textbf{3.41} & \textbf{9.55} & \textbf{3.19} & \textbf{7.53} & \textbf{2.29} & \textbf{5.45} & \textbf{3.27} & \textbf{8.63} \\
\midrule
\multirow{3}{*}{Qwen3.8-27B}
& DSpark & \textbf{5.20} & 6.74 & 4.76 & 6.07 & 4.41 & 5.65 & \textbf{5.18} & 6.85 & 4.26 & 5.79 & 3.47 & 4.53 & 3.00 & 3.82 & 4.32 & 5.64 \\
& DFlash 2 & 4.65 & 6.29 & 4.67 & 6.29 & 4.48 & 6.00 & 5.06 & 6.98 & 4.30 & 6.11 & 3.55 & 4.82 & 2.92 & 3.95 & 4.24 & 5.78 \\
& \textbf{UBTree} & 5.05 & \textbf{7.59} & \textbf{5.28} & \textbf{7.50} & \textbf{5.07} & \textbf{7.33} & 5.14 & \textbf{7.66} & \textbf{4.67} & \textbf{7.32} & \textbf{4.44} & \textbf{6.54} & \textbf{3.84} & \textbf{5.61} & \textbf{4.78} & \textbf{7.08} \\
\bottomrule
\end{tabular}
}
\endgroup
\end{table*}

\subsection{Production-scale Experiment}
\label{sec:exp-production}
\label{sec:exp-serving}

\noindent\textbf{Evaluation.}
We evaluate UBTree on three production models: Ling3-Tiny (7.9B-A1.3B MoE), Ling3-Flash (124B-A5.1B MoE) and Qwen3.8-27B~\citep{qwen38}, on the same benchmarks as in Section~\ref{sec:exp-main}. We compare UBTree with DSpark~\citep{cheng2026dspark} and DFlash~\citep{chen2026dflash} on Ling3-Tiny and Ling3-Flash, and with DSpark and DFlash~2~\citep{inco2026dflash2} on Qwen3.8-27B. We construct target-specific training sets by regenerating responses with Qwen3.8-27B under thinking-disabled decoding. For baseline and UBTree training, Qwen3.8-27B uses OpenPerfectBlend~\citep{xu2024perfect} prompts, whereas all three methods on both Ling3-Tiny and Ling3-Flash use a supervised-finetuning corpus with 2.5M samples and 15B tokens under a 64K context length. We train DSpark for 6 epochs with learning rates of $6\times10^{-4}$ on Ling3-Tiny and $2\times10^{-4}$ on Ling3-Flash. For Qwen3.8-27B, we load the officially released Qwen3.8-27B DFlash~2 checkpoint and fine-tune both the DFlash~2 selector and the UBTree selector for 2 epochs with a learning rate of $3\times10^{-4}$ on our regenerated Qwen3.8-27B rollouts. For Ling3-Tiny and Ling3-Flash, we use block size 16 for all methods. For Ling3-Flash, we train DFlash for 5 epochs with a learning rate of $6\times10^{-4}$. UBTree is then trained based on the epoch-3 DFlash checkpoint for 2 epochs, with backbone and selector learning rates of $2\times10^{-5}$ and $6\times10^{-4}$, respectively. Our tree verification uses the setup in Section~\ref{sec:exp-main}.

\noindent\textbf{SGLang Integration.}
We implement UBTree in SGLang~\citep{zheng2024sglang}. DFlash and DSpark use SGLang's official implementation, while the Qwen3.8-27B implementations follow the official DFlash~2 codebase. We use TP degree 4 on four GPUs for Ling3-Flash and Qwen3.8-27B, and TP degree 1 on one GPU for Ling3-Tiny, with the same degree for the target and draft models. All experiments use batch size 1 and greedy drafter decoding with thinking disabled. For Ling3-Tiny and Ling3-Flash, CUDA Graphs and serving optimizations are enabled for all methods. The Ling3 models and Qwen3.8-27B use hybrid attention with Kimi Delta Attention~\citep{kimi2025linear} and Gated DeltaNet~\citep{yang2024gated} layers, respectively. Inspired by Bole~\citep{wang2026bole}, each node inherits recurrent and convolution states from its parent in tree verification, and only the states and KV entries on the accepted path are committed.

\noindent\textbf{Results.}
Table~\ref{tab:production-results} shows that UBTree achieves the highest acceptance length in all 21 model--benchmark comparisons and the highest speedup in 19 of them. Averaged over the seven benchmarks, UBTree reaches $3.04\times$, $3.27\times$, and $4.78\times$ speedup on Ling3-Tiny, Ling3-Flash, and Qwen3.8-27B, respectively.

\subsection{Long-context Tasks}
\label{sec:exp-long-context}

\noindent\textbf{Setup.}
We evaluate acceptance length on Ling3-Flash using LongBench-v2~\citep{bai2024longbenchv2} for long inputs and SWE-bench~\citep{jimenez2024swebench} for long-continuation agentic tasks. For SWE-bench, all methods replay identical records. Evaluation Protocols are detailed in Appendix~\ref{app:long-context-protocol}.

\noindent\textbf{Results.}
UBTree achieves the highest acceptance length in every context-length bin, exceeding the strongest baseline by $31.0$--$39.7\%$ on LongBench-v2 and $25.6$--$29.3\%$ on SWE-bench. These results show that UBTree retains its advantage under both long initial prompts and growing agent histories. Quantitative results are available in Appendix~\ref{app:long-context}.

\begin{table*}[!th]
\centering
\caption{
Selector, training-temperature, and training-objective ablations on Qwen3-4B
with thinking disabled.
\textbf{Top:} speedup and acceptance length ($\tau$) at
$T_{\mathrm{infer}}=0$.
\textbf{Middle:} acceptance length across training and inference temperatures,
with the same checkpoints evaluated in both panels.
\textbf{Bottom:} training-objective ablation with the fixed architecture and training temperature.
}
\label{tab:ablation-selector}
\label{tab:ablation-temperature-tree}
\label{kua}

% ------------------------------------------------------------------
% Selector ablation
% ------------------------------------------------------------------
\begingroup
\small
\setlength{\tabcolsep}{2pt}
\renewcommand{\arraystretch}{0.95}
\resizebox{\textwidth}{!}{
\begin{tabular}{@{}p{4.0cm} @{\hspace{1.0em}} cc cc cc
                @{\hspace{1.0em}} cc cc cc
                @{\hspace{1.0em}} cc
                @{\hspace{1.0em}} cc@{}}
\UBTreeAblationHeader
DFlash~2 selector$+$Tree
    & 7.58 & 11.08 & 7.42 & 10.87 & 6.13 & 8.98
    & 5.88 & 8.48 & 5.64 & 8.44 & 5.40 & 7.77
    & 3.60 & 6.15 & 5.95 & 8.82 \\
UBTree selector$+$Tree
    & 8.16 & 12.11 & 7.83 & 11.67 & 6.79 & 10.06
    & 6.73 & 9.61 & 6.21 & 9.54 & 6.04 & 8.78
    & 4.26 & 7.23 & 6.57 & 9.86 \\
\quad$+$$T_{\mathrm{train}}=1$
    & \textbf{8.54} & \textbf{12.58}
    & \textbf{8.18} & \textbf{12.15}
    & \textbf{6.92} & \textbf{10.13}
    & \textbf{7.07} & \textbf{10.22}
    & \textbf{6.66} & \textbf{9.98}
    & \textbf{6.37} & \textbf{9.24}
    & \textbf{4.39} & \textbf{7.32}
    & \textbf{6.88} & \textbf{10.23} \\
\bottomrule
\end{tabular}
}
\endgroup

% ------------------------------------------------------------------
% Temperature ablation
% ------------------------------------------------------------------
\begingroup
\small
\setlength{\tabcolsep}{2pt}
\renewcommand{\arraystretch}{0.95}
\resizebox{\textwidth}{!}{
\begin{tabular}{@{}c rrrrrrrr @{\hspace{1.5em}} rrrrrrrr@{}}
\toprule
\multirow{3}{*}{$T_{\mathrm{train}}$}
& \multicolumn{8}{c}{$T_{\mathrm{infer}}=0$: Acceptance length ($\tau$)}
& \multicolumn{8}{c}{$T_{\mathrm{infer}}=1$: Acceptance length ($\tau$)} \\
\cmidrule(lr){2-9}\cmidrule(lr){10-17}
& \multicolumn{3}{c}{\textsc{Math}}
& \multicolumn{3}{c}{\textsc{Code}}
& \textsc{Chat} &
& \multicolumn{3}{c}{\textsc{Math}}
& \multicolumn{3}{c}{\textsc{Code}}
& \textsc{Chat} & \\
\cmidrule(lr){2-4}\cmidrule(lr){5-7}\cmidrule(lr){8-8}
\cmidrule(lr){10-12}\cmidrule(lr){13-15}\cmidrule(lr){16-16}
& GSM8K & M-500 & AIME & HE & MBPP & LCB & MT-B & Avg.
& GSM8K & M-500 & AIME & HE & MBPP & LCB & MT-B & Avg. \\
\midrule
0.8
    & 12.58 & 12.12 & \textbf{10.57} & 10.00 & 9.82 & 9.15 & 7.26 & 10.21
    & 11.72 & 10.47 & 7.52 & 9.39 & 9.17 & 7.63 & 6.58 & 8.93 \\
1.0
    & 12.58 & \textbf{12.15} & 10.13
    & \textbf{10.22} & \textbf{9.98} & \textbf{9.24}
    & \textbf{7.32} & \textbf{10.23}
    & \textbf{11.79} & 10.54 & \textbf{7.63}
    & 9.43 & 9.30 & 7.72 & 6.68 & \textbf{9.01} \\
1.1
    & \textbf{12.64} & 12.04 & 10.08 & 10.09 & 9.77 & 9.02 & 7.15 & 10.11
    & 11.71 & 10.52 & 7.61 & \textbf{9.48} & 9.12 & 7.71 & 6.69 & 8.98 \\
1.2
    & 12.59 & 11.94 & 9.90 & 10.10 & 9.87 & 9.01 & 7.22 & 10.09
    & 11.72 & \textbf{10.61} & 7.47 & 9.31
    & \textbf{9.40} & \textbf{7.80} & \textbf{6.72} & 9.00 \\
\bottomrule
\end{tabular}
}
\endgroup

% ------------------------------------------------------------------
% Training-objective ablation: fake two-column layout
% ------------------------------------------------------------------
\begingroup
\small
\setlength{\tabcolsep}{5pt}
\renewcommand{\arraystretch}{0.95}
\resizebox{\textwidth}{!}{
\begin{tabular}{@{}clcrr @{\hspace{3em}} clcrr@{}}
\toprule
$T_{\mathrm{infer}}$ & Selector loss & $\beta$ & Speedup & $\tau$
&
$T_{\mathrm{infer}}$ & Selector loss & $\beta$ & Speedup & $\tau$ \\
\midrule
0 & Hard-label CE & 0.1 & 6.84 & 10.08
&
1 & Hard-label CE & 0.1 & 5.99 & 8.93 \\

0 & Forward KL & 0.1 & 6.88 & 10.23
&
1 & Forward KL & 0.1 & 6.00 & 9.01 \\

0 & Forward KL & 0 & 6.87 & 10.14
&
1 & Forward KL & 0 & 6.01 & 9.01 \\
\bottomrule
\end{tabular}
}
\endgroup

\end{table*}

\subsection{Ablation Studies}
\label{sec:exp-ablation}

 \textbf{Architecture \& High-temperature Training.} We first ablate our two designs, selector architecture and high-temperature training temperature, on Qwen3-4B under $T_{\mathrm{infer}}=0$ and setups in Section~\ref{sec:exp-setup}. As shown in Table~\ref{tab:ablation-selector} (Top), our selector architecture, combining deep codebooks and depth calibration, improves both speedup and acceptance length across all benchmarks, increasing their averages from $5.95\times$ to $6.57\times$ and from $8.82$ to $9.86$. A higher training temperature further helps reach an average speedup of $6.88\times$ and acceptance length of $10.23$, validating our tree-native training.
 
 \textbf{Training Temperature Sweeping.} We then examine the optimal training temperature $T_{train}$ of UBTree on Qwen3-4B by varying $T_{\mathrm{train}}\in\{0.8,1.0,1.1,1.2\}$ and evaluating each checkpoint at $T_{\mathrm{infer}}\in\{0,1\}$ in Table~\ref{tab:ablation-temperature-tree} (Middle). Among the tested temperatures, $T_{\mathrm{train}}=1$ achieves the highest average acceptance length at both inference temperatures. 
 
 \textbf{Training Objectives.} We also compare different choices of training objectives by the average speedup and acceptance length over seven benchmarks in Table~\ref{tab:ablation-temperature-tree} (Bottom). Both using the cross-entropy (CE) loss for the selector and disabling the CE loss for the proposer degrade the acceptance length at two inference temperatures. Additional ablation details are provided in Appendix~\ref{app:ablation-protocol}.

\section{Related Work}
\label{sec:related-work}

\noindent\textbf{Parallel Drafting and Token-wise Dependencies.}
DFlash generates tokens in a draft block in parallel~\citep{chen2026dflash}. Domino and DSpark restore token-wise dependencies through sequential correction~\citep{huang2026domino,cheng2026dspark}, whereas DFlash~2 uses a low-rank selector to score adjacent candidate pairs in parallel for single-path drafting~\citep{inco2026dflash2}. UBTree retains this scoring formulation but redesigns the token representations and learns depth-dependent weights for combining proposer logits and selector scores in tree verification.

\noindent\textbf{Tree-based Speculative Decoding.}
SpecInfer and Medusa retain multiple draft paths for tree verification~\citep{miao2024specinfer,cai2024medusa}, while EAGLE-2 adapts tree structure to proposal confidence~\citep{li2024eagle2}. DART applies $n$-gram-guided pruning to parallel predictions~\citep{liu2026dart}. DDTree and DARTree construct trees from block-parallel proposals using position-wise probabilities and path-conditioned correction, respectively~\citep{ringel2026accelerating,li2026dartree}. Compared to existing tree drafting methods, UBTree exploits \textit{tree-native} training to supervise the selector with target probabilities over alternatives rather than individual sampled tokens.

\section{Conclusion}
UBTree offers a new perspective on speculative decoding: tree verification calls for designs that explicitly account for multiple plausible generation paths. Its strong performance across benchmarks demonstrates the effectiveness of this principle. We also discuss limitations of our work in Appendix~\ref{app:limitations}. Future work could further advance inference-time tree size selection, adapting the verification budget to balance candidate coverage and verification cost across different hardware, serving loads, and latency requirements.

% Custom bibliography entries only.
\bibliographystyle{antgroup}
\bibliography{custom}

\clearpage
\appendix
\section{Tree Construction Details}
\label{app:tree-construction}

UBTree follows the search schedule of DARTree's pruned variant (Algorithm~1 in \citealp{li2026dartree}), reviewed in Section~\ref{sec:prelim-tree}. DARTree evaluates path-conditioned correction probabilities and updates correction states between depths; UBTree instead supplies precomputed proposer and selector scores, so these correction-state updates are not needed.

\noindent\textbf{Path Scores.}
For a fixed verified context $x_{\le t}$, we use the candidate sets from Section~\ref{sec:method-unigram} and combined scores from~\eqref{eq:ubtree-depth}. These scores incorporate both proposer logits and selector scores conditioned on the proposer's hidden states. For each predecessor $a\in\mathcal C_{i-1}$, normalization over successor candidates defines a conditional distribution, whose log-probabilities accumulate along a candidate path:
\begin{equation}
    \begin{aligned}
        q_i^{\mathrm{bi}}(b\mid a,x_{\le t})
        &=\frac{\exp s_i(a,b)}
        {\sum_{v\in\mathcal C_i}\exp s_i(a,v)},
        \qquad b\in\mathcal C_i,\\
        \ell(y_{1:d})
        &=\sum_{i=1}^{d}
        \log q_i^{\mathrm{bi}}(y_i\mid y_{i-1},x_{\le t}).
    \end{aligned}
    \label{eq:ubtree-path}
\end{equation}
Here, $y_0=x_t$ and $1\le d\le\gamma$ is the path depth. The root represents the empty draft prefix $\epsilon$, with $\ell(\epsilon)=0$. Each non-root node is identified by its complete draft prefix $y_{1:d}$; identical tokens reached through different parents remain distinct nodes.

\noindent\textbf{Depth Bonus.}
Following DARTree's scoring form, the search permits a nonpositive depth bonus when ranking tree nodes. The reported UBTree evaluations use $\zeta=0$:
\begin{equation}
    \ell_\zeta(y_{1:d})=\ell(y_{1:d})+\zeta d,
    \qquad \zeta=0.
    \label{eq:ubtree-depth-bonus}
\end{equation}
We set $\ell_\zeta(\epsilon)=0$ and use $\zeta$ to distinguish this depth bonus from the training-loss weight $\beta$. The bonus is applied to path scores after probability normalization. It leaves rankings within each depth unchanged and penalizes deeper nodes during global pruning only when $\zeta<0$.

\noindent\textbf{Depth-Wise Expansion.}
Let $\mathcal F_d$ be the retained frontier at depth $d$, initialized by $\mathcal F_0=\{\epsilon\}$. At each depth, we extend every retained path with the candidates at that position and select the highest-scoring extensions:
\begin{equation}
    \begin{aligned}
        \mathcal E_d
        &=\{y_{1:d}:y_{1:d-1}\in\mathcal F_{d-1},\ y_d\in\mathcal C_d\},\\
        \mathcal F_d
        &=\operatorname{Top}_{W}(\mathcal E_d;\ell_\zeta),
        \qquad d=1,\ldots,\gamma.
    \end{aligned}
    \label{eq:ubtree-frontier}
\end{equation}
We interpret $y_{1:0}=\epsilon$. The operator $\operatorname{Top}_{W}(\mathcal E_d;\ell_\zeta)$ retains up to $W$ nodes with the largest depth-penalized scores across all extensions at depth $d$, not $W$ children per parent. Scoring an extension adds its conditional log-probability from~\eqref{eq:ubtree-path} and $\zeta$ to its parent's score using the precomputed combined scores, without another selector evaluation.

\noindent\textbf{Global Pruning.}
The retained frontiers form a candidate supertree. We select at most $B$ non-root nodes from this supertree by the same depth-penalized score:
\begin{equation}
    \mathcal S=\bigcup_{d=1}^{\gamma}\mathcal F_d,
    \qquad
    \mathcal T=\{\epsilon\}\cup\operatorname{Top}_{B}(\mathcal S;\ell_\zeta).
    \label{eq:ubtree-pruning}
\end{equation}
The root is kept separately and does not count toward $B$. Since every normalized probability is at most one and $\zeta\le0$, $\ell_\zeta(y_{1:d})\le\ell_\zeta(y_{1:d-1})$. Preferring ancestors when scores tie therefore ensures that every selected node retains all its ancestors, following the prefix-monotonicity argument in Lemma~1 of \citet{li2026dartree}. This selection is restricted to the materialized supertree $\mathcal S$; finite-width expansion can discard paths before global pruning.

\noindent\textbf{Construction Summary.}
  Algorithm~\ref{alg:ubtree-construction} summarizes tree construction
  from precomputed proposer and selector scores, without further
  selector evaluation. We use $y_{1:0}=\epsilon$ and $y_0=x_t$.
  The operator $\operatorname{Top}_m$ retains up to $m$
  highest-scoring nodes, preferring ancestors when scores tie.

  \begin{algorithm}[t]
  \caption{UBTree Construction from Precomputed Scores}
  \label{alg:ubtree-construction}
  \begin{algorithmic}[1]
      \Statex \textbf{Input:}
      Verified context $x_{\le t}$,
      candidate sets $\{\mathcal C_i\}_{i=1}^{\gamma}$,
      precomputed combined scores $\{s_i\}_{i=1}^{\gamma}$,
      search width $W$, and non-root node budget $B$.
      \Statex \textbf{Output:}
      A prefix-closed tree $\mathcal T$
      with at most $B$ non-root nodes.

      \State $\zeta \gets 0$
      \State Precompute all candidate-pair log-probabilities
      $\log q_i^{\mathrm{bi}}$ using~\eqref{eq:ubtree-path}.
      \State $\mathcal F_0 \gets \{\epsilon\}$,
      $\ell_\zeta(\epsilon) \gets 0$,
      $\mathcal S \gets \varnothing$

      \For{$d=1,\ldots,\gamma$}
          \State $\mathcal E_d \gets
          \{y_{1:d} :
          y_{1:d-1}\in\mathcal F_{d-1},\
          y_d\in\mathcal C_d\}$
          \ForAll{$y_{1:d}\in\mathcal E_d$}
              \State $\ell_\zeta(y_{1:d}) \gets
              \ell_\zeta(y_{1:d-1})
              + \log q_d^{\mathrm{bi}}
              (y_d\mid y_{d-1},x_{\le t})
              + \zeta$
          \EndFor
          \State $\mathcal F_d \gets
          \operatorname{Top}_{W}(\mathcal E_d;\ell_\zeta)$
          \State $\mathcal S \gets
          \mathcal S \cup \mathcal F_d$
      \EndFor

      \State $\mathcal T \gets
      \{\epsilon\}\cup
      \operatorname{Top}_{B}(\mathcal S;\ell_\zeta)$
      \State \Return $\mathcal T$
  \end{algorithmic}
  \end{algorithm}

\noindent\textbf{Verification via Tree-attention.}
After tree construction, we verify the prefix-closed draft tree with at most \(B=64\) non-root nodes in a single target-model forward pass using tree attention. Each node is evaluated under its corresponding autoregressive prefix, and verification follows the child matching the target-model sample at each step until the first mismatch. The first unmatched sample is retained as the bonus token. After verification, the KV cache and recurrent states along the accepted path are carried forward to the next decoding round, while those of all other branches are discarded. Since every committed token is sampled from the target model under exactly the same prefix as in standard autoregressive decoding, the procedure is lossless.
\section{Experimental Setup Details}
\label{app:experimental-setup}

We provide additional training and evaluation settings for the academic-scale experiments, long-context evaluations, and ablations in Section~\ref{sec:experiments}.

\subsection{Training and Inference Configuration}
\label{app:training-configuration}

Table~\ref{tab:ubtree-configuration} summarizes the configuration for the academic-scale experiments. We initialize the depth-calibration parameters to $\rho_i=\kappa_i=0$, so the proposer logits and selector scores initially have unit weights. Academic-scale inference loads the target and drafter weights in BF16 and captures draft post-processing in a single CUDA Graph. For Ling3-Tiny and Ling3-Flash, UBTree uses 49 and 32 total verification nodes including the anchor, respectively.

\begin{table}[t]
\centering
\caption{UBTree configuration for the academic-scale experiments, unless otherwise specified. The verification budget excludes the anchor.}
\label{tab:ubtree-configuration}
\begin{tabular}{ll}
\toprule
Setting & Value \\
\midrule
Proposer training & 6 epochs \\
Joint proposer--selector training & 2 epochs \\
Selector rank & 256 \\
Codebook MLP hidden width & 1024 \\
Training candidate support size & 256 \\
\midrule
Draft depth & 15 future positions plus an anchor \\
Candidate tokens per position $K$ & 64 \\
Search width $W$ & 12 \\
Verification budget $B$ & 64 non-root nodes \\
Depth bonus $\zeta$ & $0$ \\
\bottomrule
\end{tabular}
\end{table}

\subsection{Evaluation and Timing Protocol}
\label{app:evaluation-protocol}

For our academic-scale evaluations, we use fixed seed-0 subsets with the sample counts listed in Table~\ref{tab:evaluation-samples}. Each configuration is evaluated once on the selected subset. Generation stops at EOS or after 2,048 new tokens. MT-Bench contains 80 two-turn dialogues, yielding 160 evaluated turns.

For each question or dialogue turn, time per output token (TPOT) is computed from the elapsed decoding time and the output-token count within the same timing window. The same decoding-time measurement protocol is used for all locally evaluated academic-scale methods and the autoregressive baseline. Within each benchmark, speedup is the ratio of the mean autoregressive TPOT to the mean speculative-decoding TPOT. Acceptance length is averaged over questions or dialogue turns.

\begin{table}[t]
\centering
\caption{Evaluation subset sizes. MT-Bench contains two-turn dialogues; the other counts refer to individual examples.}
\label{tab:evaluation-samples}
\begin{tabular}{lr}
\toprule
Benchmark & Examples / dialogues \\
\midrule
GSM8K & 128 \\
MATH-500 & 128 \\
AIME & 30 \\
HumanEval & 128 \\
MBPP & 128 \\
LiveCodeBench & 128 \\
MT-Bench & 80 \\
\bottomrule
\end{tabular}
\end{table}

\subsection{Long-context Evaluation Protocol}
\label{app:long-context-protocol}

\noindent\textbf{Shared Settings.}
We compare UBTree with DSpark and DFlash on Ling3-Flash using tensor parallelism of degree 4 on four GPUs, greedy decoding, and thinking disabled. Both long-context evaluations use concurrency 1 with decode CUDA Graphs enabled. We measure root-inclusive acceptance length $\tau$ as the number of generated tokens per target verification call, including the target-produced token. All methods use a maximum draft depth of 15, giving a maximum acceptance length of 16. DSpark and DFlash verify a single chain, while UBTree uses a budget of 31 non-root tree nodes per round.

\noindent\textbf{LongBench-v2.}
We evaluate long-input generation on LongBench-v2~\citep{bai2024longbenchv2}. From the 503 source examples, we retain the same 401 requests with at most 260K prompt tokens for all methods. Generation stops at EOS or after 2,048 new tokens. We group requests by their prompt length before generation into six context-length bins with upper boundaries of 32K, 64K, 96K, 128K, 192K, and 260K tokens. Each bin contains the naturally occurring benchmark inputs in that length range. Table~\ref{tab:long-context-longbench} reports acceptance length and request counts for these bins, together with results over the full evaluation subset.

\noindent\textbf{SWE-bench.}
We evaluate acceptance as agent histories grow on SWE-bench~\citep{jimenez2024swebench}. We first run target-only inference on 30 cases and freeze the resulting 1,482 complete requests. Each method then replays the identical recorded messages, tool calls, and tool outputs with a fixed generation limit of 256 tokens. This shared replay protocol aligns the agent context across methods at every request. To match the context-length axis in Figure~\ref{fig:ubtree-teaser}(c), we group requests by the increase in prompt tokens relative to the first request of the corresponding case, using bins $\le0$, 0--4K, 4--8K, 8--16K, 16--32K, 32--64K, and 64--131K. The $\le0$ bin contains requests whose prompt length is no greater than the initial prompt, including context truncation or reconstruction events. Table~\ref{tab:long-context-swe} reports acceptance length and the number of cases represented in each bin; its Overall column reports request-pooled acceptance length over all 1,482 requests.

As a complementary analysis, we summarize the same replay requests by trajectory progress. We divide each case chronologically into four stages containing approximately equal numbers of requests: 0--25\%, 25--50\%, 50--75\%, and 75--100\%. Within each case and stage, we compute acceptance length as total generated tokens divided by total target verification calls, then average equally across the 30 cases. These stages measure relative progress through an agent trajectory. Since the amount of context added per request varies across cases and requests, the stage boundaries do not correspond to fixed context lengths.

\subsection{Ablation Protocol}
\label{app:ablation-protocol}

For the training-temperature ablation in Table~\ref{tab:ablation-temperature-tree}, we regenerate target responses at each $T_{\mathrm{train}}$ and use them for both proposer training and joint proposer--selector training. Each joint run starts from the proposer trained at the corresponding temperature. The target distribution for KL supervision uses the same $T_{\mathrm{train}}$, following Eq.~\eqref{eq:ubtree-training-distributions}. Across temperatures, we retain the same selector architecture, training configuration, and tree-search configuration. Each resulting checkpoint is evaluated at both $T_{\mathrm{infer}}=0$ and $T_{\mathrm{infer}}=1$. For the training-objective ablation, all variants start from the same proposer checkpoint and use identical training responses. Hard-label selector CE excludes positions whose sampled target token falls outside the training candidate support, whereas forward KL supervises all valid positions.

\FloatBarrier

\section{Additional Experiments}
\label{app:additional-experiments}

\subsection{Drafting Latency Analysis}
\label{sec:exp-latency}

 We profile Qwen3-4B using Hugging Face Transformers-based implementations on 128 GSM8K questions at $T_{\mathrm{infer}}=0$, using the search budgets in Section~\ref{sec:exp-setup}. For each question, all methods run on the same GPU in randomized order, and we average measurements over ten complete passes. Timing excludes prefill and the first complete speculative round. Stage latencies are measured with CUDA events, while total round latency is measured independently using synchronized wall-clock time. For UBTree, draft post-processing includes LM-head projection, candidate selection, selector scoring, and tree construction, which are timed jointly. The ms/token column reports TPOT averaged over questions.

Table~\ref{tab:latency-breakdown} shows that UBTree reduces draft post-processing latency from DARTree's 3.36 ms to 1.38 ms per round, a reduction of 58.8\%, while target verification takes approximately 20.9 ms for both methods. The principal latency reduction therefore occurs on the parallel selector side. UBTree achieves a higher acceptance length than DARTree (12.58 versus 12.50) while reducing decoding latency from 2.41 to 2.21 ms per token.

\begin{table*}[htbp]
\centering
\caption{Drafting Latency profiling on Qwen3-4B. Stage and total latencies are in ms per verification round.}
\label{tab:latency-breakdown}
\small
\resizebox{\textwidth}{!}{
\begin{tabular}{lrrrrrrr}
\toprule
Method & Proposer stage & Draft post-processing & Verification & Other & Total & $\tau$ & ms/token \\
\midrule
DARTree & 3.51 & 3.36 & 20.94 & 1.64 & 29.47 & 12.50 & 2.41 \\
UBTree & 3.41 & 1.38 & 20.88 & 1.61 & 27.30 & 12.58 & 2.21 \\
\bottomrule
\end{tabular}
}
\vspace{2pt}
\end{table*}

\subsection{Long-context Tasks}
\label{app:long-context}

Tables~\ref{tab:long-context-longbench} and~\ref{tab:long-context-swe} report the quantitative results, together with the overall acceptance length on each benchmark. All results follow the long-context evaluation protocol in Appendix~\ref{app:long-context-protocol}.

\begin{table*}[!t]
\centering
\caption{Acceptance length ($\tau$) on LongBench-v2 with Ling3-Flash, grouped by request-start prompt length. All methods use the same 401 requests. The Overall column reports results over the full evaluation subset. Bold marks the best acceptance length per column.}
\label{tab:long-context-longbench}
\begingroup
\small
\setlength{\tabcolsep}{4pt}
\renewcommand{\arraystretch}{0.95}
\resizebox{\textwidth}{!}{
\begin{tabular}{@{}l ccccccc@{}}
\toprule
Prompt tokens & $\le32$K & 32--64K & 64--96K & 96--128K & 128--192K & 192--260K & Overall \\
Requests & 109 & 71 & 47 & 69 & 73 & 32 & 401 \\
\midrule
DSpark & 4.889 & 4.970 & 4.281 & 4.286 & 4.642 & 4.617 & 4.676 \\
DFlash & 4.459 & 4.503 & 3.831 & 4.025 & 4.298 & 4.320 & 4.272 \\
\textbf{UBTree} & \textbf{6.454} & \textbf{6.752} & \textbf{5.609} & \textbf{5.799} & \textbf{6.485} & \textbf{6.434} & \textbf{6.300} \\
\bottomrule
\end{tabular}
}
\endgroup
\end{table*}

\begin{table*}[!t]
\centering
\caption{Acceptance length ($\tau$) on SWE-bench with Ling3-Flash, grouped by accumulated prompt tokens relative to the first request in each case. Covered cases gives the number of cases represented in each bin. The Overall column reports request-pooled acceptance length over all 1,482 requests from 30 cases. Bold marks the best acceptance length per column.}
\label{tab:long-context-swe}
\begingroup
\small
\setlength{\tabcolsep}{4pt}
\renewcommand{\arraystretch}{0.95}
\resizebox{\textwidth}{!}{
\begin{tabular}{@{}l cccccccc@{}}
\toprule
Accumulated tokens & $\le0$ & 0--4K & 4--8K & 8--16K & 16--32K & 32--64K & 64--131K & Overall \\
Covered cases & 30 & 30 & 28 & 27 & 23 & 13 & 2 & 30 \\
\midrule
DSpark & 3.793 & 3.899 & 4.330 & 4.270 & 4.430 & 4.412 & 4.473 & 4.277 \\
DFlash & 3.389 & 3.436 & 3.963 & 3.794 & 4.044 & 3.923 & 4.023 & 3.826 \\
\textbf{UBTree} & \textbf{4.794} & \textbf{5.041} & \textbf{5.449} & \textbf{5.481} & \textbf{5.714} & \textbf{5.544} & \textbf{5.616} & \textbf{5.450} \\
\bottomrule
\end{tabular}
}
\endgroup
\end{table*}

\noindent\textbf{Results.}
UBTree achieves the highest acceptance length in every context-length bin on both benchmarks. On LongBench-v2, UBTree reaches $\tau=5.609$--$6.752$, improving over the strongest baseline, DSpark, by $31.0$--$39.7\%$. At 192--260K prompt tokens, UBTree attains $\tau=6.434$, compared with 4.617 for DSpark and 4.320 for DFlash, retaining nearly the same acceptance length as in the shortest bin ($6.454$). Overall, UBTree achieves $\tau=6.300$, a $34.7\%$ improvement over DSpark. On SWE-bench, UBTree reaches $\tau=4.794$--$5.714$ across accumulated-context bins, exceeding DSpark by $25.6$--$29.3\%$. Its overall request-pooled acceptance length is $5.450$, compared with $4.277$ for DSpark and $3.826$ for DFlash, improving over the strongest baseline by $27.4\%$. These results demonstrate UBTree's sustained acceptance advantage across both long initial prompts and growing agent histories.

\noindent\textbf{Trajectory-stage Results.}
Under the complementary trajectory-stage aggregation defined in Appendix~\ref{app:long-context-protocol}, UBTree achieves acceptance lengths of 5.116, 5.563, 5.448, and 5.644 across the four successive SWE-bench stages, compared with 3.942, 4.321, 4.241, and 4.461 for the strongest baseline, DSpark. The corresponding gains range from $26.5\%$ to $29.8\%$. This summary captures UBTree's advantage throughout agent trajectories, while the $25.6$--$29.3\%$ gains reported in the main text correspond to the accumulated-context results in Table~\ref{tab:long-context-swe}. The two ranges summarize the same replay workload under different groupings.

\FloatBarrier

\section{Limitations}
\label{app:limitations}
Despite UBTree's strong empirical performance, two limitations remain.

\noindent\textit{Tree drafting is sensitive to very high concurrency.}
Speculative decoding is motivated by the spare compute available when autoregressive inference is limited by memory bandwidth, as is common in LLM decoding. Verifying multiple draft tokens together uses this capacity to amortize memory-access costs. As concurrency increases and inference becomes compute-bound, the spare capacity available for speculation shrinks, potentially eliminating speedups for both single-path and tree-based methods. Sensitivity to this transition is shared by all tree-based drafters and is not specific to UBTree. Tree verification spends additional compute on alternative continuations and therefore favors the part of the compute--bandwidth spectrum with more spare compute. Adaptive tree drafting can mitigate this limitation by varying the verification budget with available compute: branching allows the number of verified candidates to vary beyond the token count of a fixed single-path block, although the proposer block size still bounds draft depth. Our load-dependent tree budgets already retain speedups at the tested concurrency levels (Table~\ref{tab:concurrency-results}), and more flexible runtime adaptation could further improve resource use. UBTree can also serve low-concurrency workloads where spare compute is abundant. We therefore view this limitation as a constraint on the favorable deployment regime, rather than a fundamental obstacle to UBTree's practical value.

\noindent\textit{Adaptive tree strategies rely on searched configurations.}
Our current adaptive strategy uses tree budgets $B$ and search widths $W$ selected through configuration search for each model and concurrency level. These settings are fixed for a given serving load; they do not determine the budget from the transition-score distribution of each decoding round. Although the tree's contents depend on the predicted scores, the budget-selection policy remains empirically tuned and is not guaranteed to be optimal as prediction uncertainty and available compute change. A more principled strategy would select tree sizes at each decoding round using calibrated estimates of acceptance benefit derived from transition scores, together with a cost model reflecting hardware and serving load. We leave developing and evaluating such a policy to future work.

\section{Additional Related Work}
\label{app:additional-related-work}

\noindent\textbf{Speculative Decoding.} In addition to autoregressive~\citep{leviathan2023fast,chen2023accelerating,li2024eagle}, parallel~\citep{chen2026dflash,liu2026dart,inco2026dflash2}, semi-autoregressive~\citep{huang2026domino,cheng2026dspark}, and tree-based~\citep{cai2024medusa,miao2024specinfer,li2024eagle2,li2026eagle,ringel2026accelerating,li2026dartree} drafting methods mentioned, existing research also explored mixing methods for speculative decoding. PARD adapts autoregressive models for parallel token prediction~\citep{an2025pard}, while DiffuSpec uses pretrained diffusion language models as drafters~\citep{li2026diffuspec}.

\noindent\textbf{Lossy Acceleration for Large Language Models.}
Lossy acceleration reduces inference cost by allowing deviations from the target model's output distribution. BiLD~\citep{kim2023biglittle} coordinates small and large models through confidence-based fallback and discrepancy-based rollback. The lossy extension of DistillSpec~\citep{zhou2024distillspec} relaxes token acceptance using a lenience factor, while Fuzzy Speculative Decoding~\citep{holsman2025fuzzy} bases acceptance on divergence between draft and target distributions. Speculative cascades~\citep{narasimhan2025faster} implement model-deferral rules through speculative execution, selecting between draft and target distributions; DIVERSED~\citep{wang2026diversed} instead learns context-dependent mixtures of these distributions for verification. Judge Decoding~\citep{bachmann2025judge,garipov2026autojudge} accepts otherwise-rejected tokens using a lightweight correctness classifier on target hidden states. In contrast to these relaxed verification methods, UBTree improves candidate coverage and path quality through joint proposer--selector training while retaining target-distribution-preserving verification.

\noindent\textbf{Multi-token Language Models for Fast Generation.}
Another line of work incorporates multi-token prediction into the language model itself. \citet{gloeckle2024better} train multiple prediction heads on a shared backbone to predict several future tokens simultaneously. Diffusion language models, including LLaDA~\citep{nie2025large} and Dream~\citep{xie2025dream}, learn to recover masked tokens and generate text through iterative parallel denoising. LLaDA is trained from scratch, whereas Dream adapts pretrained autoregressive weights. Block diffusion~\citep{arriola2025block} combines autoregressive generation across blocks with parallel denoising within each block, enabling flexible-length generation and prefix KV caching. Large-scale block diffusion with different recipes enriches the explored space of taming diffusion for multi-token fast generation, including WeDLM~\citep{liu2025wedlm}, Mercury~\citep{inception2025mercury}, and DiffusionGemma~\citep{diffusiongemma2026report}. These approaches modify the generative model or its training to enable multi-token generation, while having the risk of degrading the sample quality.

\end{document}